\documentclass[runningheads]{llncs}
\usepackage{graphicx}
\usepackage{amsmath}
\usepackage[noend]{algpseudocode}
\usepackage{algorithmicx,algorithm}
\usepackage{diagbox}
\usepackage{booktabs}
\usepackage{multirow}
\begin{document}
%
\title{CADM: Confusion Model-based Detection Method for Real-drift in Chunk Data Stream}
\titlerunning{{CADM: Confusion and Detection Method}}

\author{}
\author{Songqiao Hu\inst{1} \and
Zeyi Liu\inst{2} \and
Xiao He\inst{2}}
\authorrunning{Songqiao Hu {\emph et al.}}
%
\institute{School of Automation, Beijing Institute of Technology, Beijing 100081, China \email{1120193091@bit.edu.cn} \and
Department of Automation, Tsinghua University, Beijing 100084, China
\email{liuzy21@mails.tsinghua.edu.cn}\\
\email{hexiao@tsinghua.edu.cn}}

\maketitle              
\begin{abstract}
Concept drift detection has attracted considerable attention due to its importance 
in many real-world applications such as health monitoring and fault diagnosis. Conventionally, most advanced approaches will be of poor performance when the evaluation criteria of the environment has changed (i.e. concept drift), either can only detect and adapt to virtual drift. In this paper, we propose a new approach to detect real-drift in the chunk data stream with limited annotations based on concept confusion. When a new data chunk arrives, we use both real labels and pseudo labels to update the model after prediction and drift detection. In this context, the model will be confused and yields prediction difference once drift occurs. We then adopt cosine similarity to measure the difference. And an adaptive threshold method is proposed to find the abnormal value. Experiments show that our method has a low false alarm rate and false negative rate with the utilization of different classifiers.
\end{abstract}
\section{Introduction}
Concept drift is a nonnegligible factor in data analysis of dynamic systems. For example, much data is collected by sensors, but the sensors' output is vulnerable to its structure and the surrounding environment, of which temperature is the most influential. Under different temperatures, the distribution range of the collected data and even their categories will change. If the monitor system fails to detect the change, it will be difficult to provide correct decision-making suggestions when risks happen \cite{DBLP:journals/tim/LiuDZDH21}.

In the literature, several advanced studies have been proposed for solving this problem.  Gama {\emph et al.} \cite{DDM} proposed Drift Detection Method (DDM) to detect whether the overall online error rate increased greatly, which was used to judge whether the warning level or the drift level was reached. \cite{ADWIN} proposed a two-time window-based drift detection algorithm named ADaptive WINdowing (ADWIN). ADWIN indicates that concept drift occurs if the means of data in two windows differ significantly. In \cite{CPSSDS}, conformal prediction was introduced to detect concept drift. Concept drift was confirmed if the conformal prediction result in the two data chunks showed a great difference. Otherwise, pseudo labels were used to update the model. Lu {\emph et al.} \cite{Adaptive_chunksize} adopted ensemble learning and adjusted each base classifier's weight according to their performance. Once a base classifier was dropped out and a new classifier needed to generate, the size of the train data chunk increased continuously until the variance stopped rising. Using the selected size data chunk to train a new base classifier could adapt to concept drift. DSPOT proposed by Siffer {\emph et al.} \cite{Extreme} believed that anomalies are usually extreme values and utilized extreme theory to fit the extreme value distributions for calculating the dynamic concept drift threshold. Sethi {\emph et al.} \cite{GC3} proposed the GC3 framework, which uses grid density clustering and a unified grid density sampling mechanism to achieve better performance with a lower label rate. The concept drift was then detected, which is the idea of feedback.

Generally, there are two main types of concept drift: real concept drift changes in $P(y|{\textbf x})$, and virtual concept drift changes in $P({\textbf x})$. Nevertheless, many of the above methods can only detect virtual drift but do nothing for real drift. \cite{ADWIN} only considers the mean of the data but is unconcerned about the labels of the data. So it can just handle virtual drift in principle. \cite{CPSSDS} seems to be put forward for real concept drift. However, judging whether two data chunks have significant distribution differences only uses the pseudo labels predicted by the same classifier. Therefore, it cannot find the change in the labels. In \cite{Extreme}, only the case of one-dimensional time series is taken into account, and whether it is an outlier is given according to the value of the feature, and the samples do not have labels at all. So the method is similar to judging concept drift from {\emph P}({\bfseries $x$}). Another typical problem in data mining and concept drift is label cost. In real situations, obtaining actual data labels usually requires a lot of time and energy, especially in the case of a high-speed data stream. Therefore, it is unrealistic to regard the concept drift as a supervised learning problem \cite{DDM}\cite{Adaptive_chunksize}\cite{liu2022OABL}. \cite{GC3} is one of the few methods that consider both actual drift and labeling cost. The main framework is based on the idea of feedback. Namely, some samples are selected for annotations after the prediction. Then, the classifier is adjusted according to the difference between the prediction and labeling results.

In this paper, we propose a novel chunk-based confusion model method called \textbf{C}onfusion \textbf{A}nd \textbf{D}etection \textbf{M}ethod (CADM) to deal with real concept drift with limited annotations. 
The main contribution of this paper can be summarized as follows: (1) A novel method is provided to detect real concept drift with limited labels in chunk data streams; (2) To the best of our knowledge, it is the first method that requires labels after concept drift detection. The drift detection results can then be used to guide the annotation costs; (3) The effectiveness is verified based on numerous experiments.

The rest of the paper is organized as follows. Section 2 introduces the method and datasets adopted in this study. Section 3 describes and discusses some experimental results and section 4 presents the conclusions.

\section{Proposed Method}


\subsection{Motivation}

\begin{figure}[htbp]
    \begin{minipage}[t]{0.5\linewidth}
        \centering
        \includegraphics[width=\textwidth]{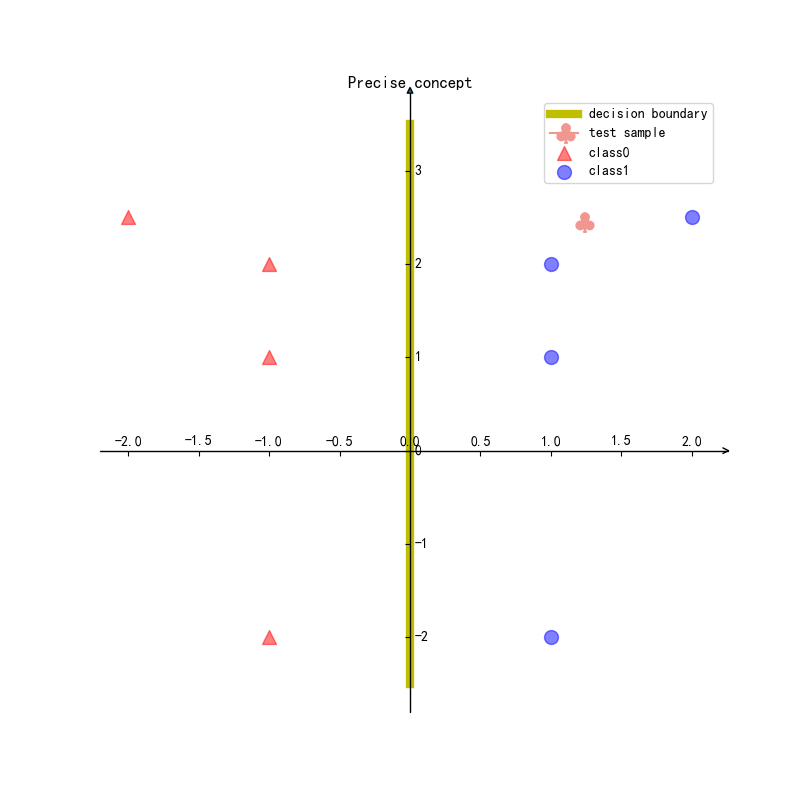}
        \centerline{(a) PC = [4.55$\times$$10^{-6}$, 0.999995]}
    \end{minipage}%
    \begin{minipage}[t]{0.5\linewidth}
        \centering
        \includegraphics[width=\textwidth]{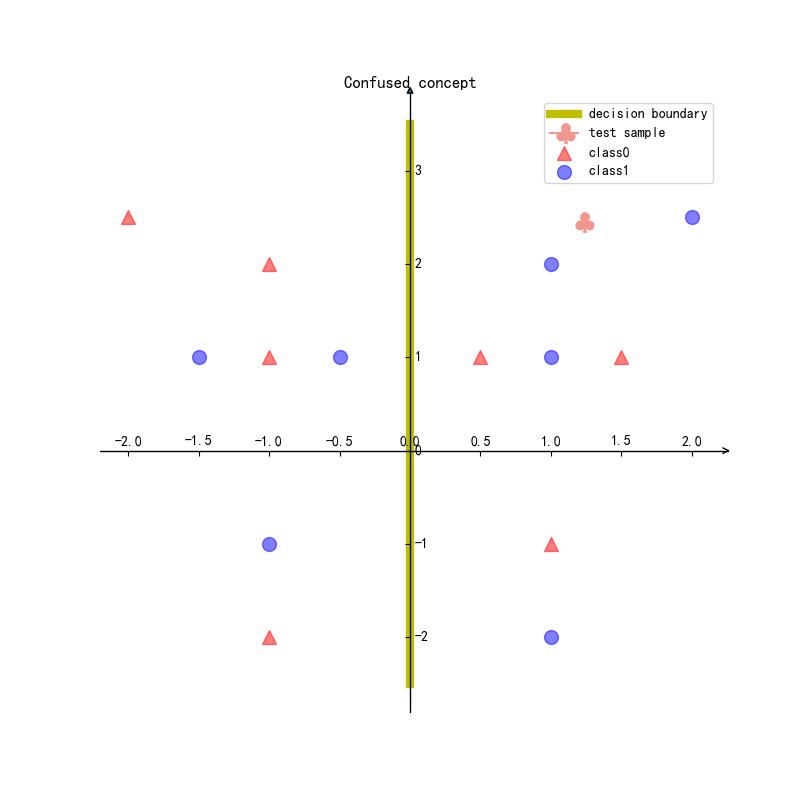}
        \centerline{(b) PC = [0.3954, 0.6046]}
    \end{minipage}
    \caption{The prediction confidence (PC) for different training data by Naive Bayes model. (a) The training data is from the same concept. PC is then completely certain. (b) Add another batch of data with a different concept on the basis of (a). PC becomes uncertain. Such a model is called the confusion model.}\label{Fig_1}
\end{figure}

The main idea is generated from the confusion model, which refers to a model that has a low degree of certainty in the classification judgment of samples. Generally, it can be caused by different concepts contained in training data. An example can be shown\textbf{} in Fig. \ref{Fig_1}.

In general, incremental learning models \cite{A_review} have greater advantages to serve as the confusion model compared to common models due to the ability to update the model without storing historical data. In the case of the chunk data stream, the model can predict and select some samples to update when each data chunk arrives. To detect drift and improve the prediction performance of the model, we provide some samples with hard pseudo labels according to the prediction and annotate some samples manually. The purpose of pseudo labels is to increase training samples, improve prediction performance and adapt to virtual drift. While the purpose of manual annotations is to confuse the model when the real drift occurs, which makes the difference in prediction. 

\subsection{Difference Measurement}

Difference measurement aims to measure the prediction difference in the new data chunk after the incremental update. Let {\itshape U}{$_0$=\{{\itshape x}$_1$, {\itshape x}$_2$, ..., {\itshape x}$_n$\} be the set of {\itshape n} unlabeled samples in new chunk, {\itshape h$_{t-1}$} denote the model before update, {\itshape h$_t$} represent the model after update, {\itshape h}($\cdot$)=[{\itshape p}$_1$, ..., {\itshape p}$_m$$]^T$ be the confidence probability vector for $m$ classes. In this case, we define the matrix {\itshape H}$_{t-1}$=[{\itshape h}$_{t-1}$({\itshape x}$_1$), ..., {\itshape h}$_{t-1}$({\itshape x}$_n$)}] and {\itshape H}$_t$=[{\itshape h}$_t$({\itshape x}$_1$), ..., {\itshape h}$_t$({\itshape x}$_n$)] as shown in Eqs.(1) and (2).
\begin{equation}
H_{t-1}=[h_{t-1}(x_1), ..., h_{t-1}(x_n)]=[\alpha_{t-1,1}^T, ..., \alpha_{t-1,m}^T]^T 
\end{equation}

\begin{equation}
H_t=[h_t(x_1), ..., h_t(x_n)]=[\alpha_{t,1}^T, ..., \alpha_{t,m}^T]^T
\end{equation}

In this context, the function $Sim(\cdot)$ can be defined with the utilization of model similarity as:
\begin{equation}
Sim(H_{t-1}, H_t)=\frac{1}{m}\sum_{i=1}^m\frac{\alpha_{t-1,i}^T\cdot\alpha_{t,i}}{||\alpha_{t-1,i}||\cdot||\alpha_{t,i}||}.
\end{equation}

In Eq. (3), the cosine similarity between $\alpha$$_{t-1,i}$ and $\alpha$$_{t,i}$ is calculated for each class. The mean of cosine similarity of $m$ classes is used to denote the cosine similarity of two models. Clearly, {\itshape Sim}({\itshape h}$_{t-1}$, {\itshape h}$_t$) $\in$ [-1,1] is always valid. The smaller the value is from 1, the more different the predictions of the two models are, which indicates that more likely concept drift occurs.

 \subsection{Adaptive Threshold}
 For drifting samples of the same number, the difference degree 
 of the models after updating varies with the number of samples 
 that have been trained. The more samples have been trained, the smaller the impact of new samples on the model. Therefore, it is necessary to adopt the adaptive threshold.

 Given that the model changes greatly when updates at the beginning, the oscillation of cosine similarity will be large at first and then gradually decrease.
 To this end, we propose \emph{Deviation-Adaptive Threshold} (DAT) which considers the statistical features such as mean and variance. A fixed-size window is adopted to store latest cosine similarity values. At each time stamp, the current average level of cosine similarity is judged according to the mean of the values in the window. The current oscillation level can then be obtained according to the standard deviation. 
 Then the abnormal cosine similarity value can be obtained with such an average level and oscillation level. To understand the whole procedure, the pseudo-code is shown
 in Algorithm 1.

\begin{algorithm}[t]
\caption{{\bfseries DAT} (Deviation-Adaptive Threshold)} 
\hspace*{0.02in} {\bf Input:} 
New data {\bfseries C$_{new}$}, previous window {\bfseries W$_{pre}$}, 
the size of window {\itshape {\bfseries l}}, deviation \hspace*{4.5 em}coefficient {\bfseries {\itshape k}}.\\
\hspace*{0.02in} {\bf Output:} 
Threshold {\bfseries t}, new window {\bfseries W$_{new}$}.
\begin{algorithmic}[1]
\If{length({\bfseries W$_{pre}$}) $< ${\itshape {\bfseries l}}} 
    \State Add {\bfseries C$_{new}$}$\rightarrow${\bfseries W$_{pre}$}
\Else
    \State Pop({\bfseries W$_{pre}$},0)  \hspace*{5 em}  //delete the first element
    \State Add {\bfseries C$_{new}$}$\rightarrow${\bfseries W$_{pre}$}
\EndIf
\State {\bfseries W$_{pre}$}$\rightarrow${\bfseries W$_{new}$}
\State mean({\bfseries W$_{new}$})$\rightarrow${\bfseries $\overline{x}$}
\State std({\bfseries W$_{new}$})$\rightarrow${\bfseries $\sigma$} 
\State $\overline{x}$ - {\bfseries {\itshape k}}$\sigma$$\rightarrow${\bfseries t}\\
\Return {\bfseries t}, {\bfseries W$_{new}$}
\end{algorithmic}
\end{algorithm}

\subsection{Drift Detection}
When the new data chunk arrives, it is needed to calculate the cosine similarity between the model updated by the last data chunk and the non-updated model. The window and threshold can then be updated. If the cosine similarity is lower than the threshold, the drift can then be judged to occur. The details of the algorithm can then be summarized in Algorithm 2. And the overall flow chart is summarized as shown in Fig. 2.

\begin{figure}[htbp]
\centering
\includegraphics[width=\textwidth]{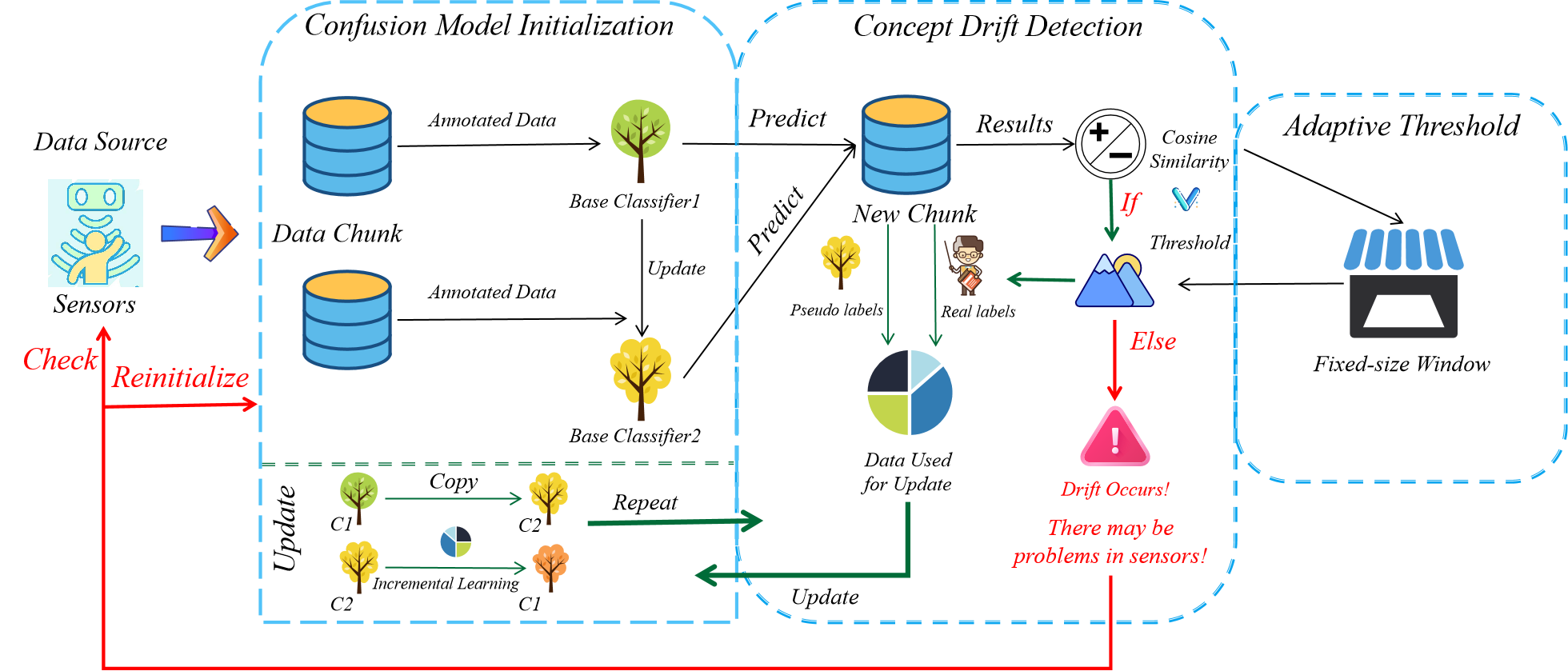}
\caption{The overall flow chart.}\label{Fig_2}
\end{figure}

 \begin{algorithm}[t]
\caption{{\bfseries CADM} (Confusion And Detection Method)} 
\hspace*{0.02in} 
{\bf Input:} 
Data chunk {\bfseries \itshape stream}=\{{\itshape D$_i$}, i=1,...,N,{\itshape D$_i$} = \{{\itshape \bfseries x$_j$}$\in$$\mathcal{X}$, j=1,...,$|${\itshape D$_i$}$|$\}\}, 
classfier type \hspace*{4.2 em}$\mathcal{C}$, label ratio {\bfseries $\lambda$}, the size of window {\itshape {\bfseries l}}, deviation coefficient {\bfseries {\itshape k}}.\\
\hspace*{0.02in} 
{\bf Output:} 
Drift points list {\bfseries \itshape L$_{drift}$}.
\begin{algorithmic}[1]
\State //Confusion model initialization.
\State $\mathcal{C}$$_1$, $\mathcal{C}$$_2$ = $\mathcal{C}$()
\State {\itshape D} = {\itshape \bfseries stream}.next\_chunk()
\State {\itshape\bfseries x$_{labeled}$} = Random({\itshape D}, {$\lambda$}$\cdot$$|${\itshape D}$|$)
\State {\itshape\bfseries y$_{labeled}$} = Give\_label({\itshape\bfseries x$_{labeled}$})
\State $\mathcal{C}$$_1$.fit({\itshape\bfseries x$_{labeled}$}, {\itshape\bfseries y$_{labeled}$})
\State $\mathcal{C}$$_2$.fit({\itshape\bfseries x$_{labeled}$}, {\itshape\bfseries y$_{labeled}$})
\State {\itshape \bfseries W} = [ ]
\State {\itshape \bfseries L$_{drift}$} = [ ]
\While{({\itshape D} = {\itshape \bfseries stream}.next\_chunk())$\neq$$\emptyset$}
    \State {\itshape \bfseries H$_{t-1}$} = $\mathcal{C}$$_1$.predict\_prob({\itshape D}) 
    \State {\itshape \bfseries H$_t$} = $\mathcal{C}$$_2$.predict\_prob({\itshape D}) 
    \State {\itshape\bfseries cos} = {\itshape Sim}({\itshape H$_{t-1}$},{\itshape H$_t$})
    \State {\itshape \bfseries t, W} = {\bfseries DAT}({\itshape\bfseries cos}, {\itshape\bfseries W}, {\bfseries\itshape l}, {\itshape\bfseries k}) \hspace*{0.5 em} //calculate the threshold and update the window
    \If{{\itshape\bfseries cos} $< ${\itshape\bfseries t}}
        \State Add chunk\_index $\rightarrow$ {\itshape \bfseries L$_{drift}$}
        \State {\itshape \bfseries W} = [ ]
        \State $\mathcal{C}$$_1$, $\mathcal{C}$$_2$ = $\mathcal{C}$()
        \State {\itshape\bfseries x$_{labeled}$} = Random({\itshape D}, {$\lambda$}$\cdot$$|${\itshape D}$|$)
        \State {\itshape\bfseries y$_{labeled}$} = Give\_label({\itshape\bfseries x$_{labeled}$})
        \State $\mathcal{C}$$_1$.fit({\itshape\bfseries x$_{labeled}$}, {\itshape\bfseries y$_{labeled}$})
        \State $\mathcal{C}$$_2$.fit({\itshape\bfseries x$_{labeled}$}, {\itshape\bfseries y$_{labeled}$}) \hspace*{4.2 em}//drift occurs and reinitialize the models
    \Else
        \State {\itshape\bfseries x$_{labeled}$} = Random({\itshape D}, $\lambda$$\cdot$$|${\itshape D}$|$)
        \State {\itshape\bfseries y$_{labeled}$} = Give\_label({\itshape\bfseries x$_{labeled}$})
        \State {\itshape\bfseries x$_{pseudo}$} = Random({\itshape D}$-$\{{\itshape\bfseries x$_{labeled}$}\}, $\lambda$$\cdot$$|${\itshape D}$|$)
        \State {\itshape\bfseries y$_{pseudo}$} = Hard\_pseudo\_label({\itshape\bfseries x$_{pseudo}$}, $\mathcal{C}$$_2$)
        \State $\mathcal{C}$$_1$ = $\mathcal{C}$$_2$
        \State $\mathcal{C}$$_2$.partial\_fit([{\itshape\bfseries x$_{labeled}$}, {\itshape\bfseries x$_{pseudo}$}], [{\itshape\bfseries y$_{labeled}$},{\itshape\bfseries y$_{pseudo}$}])
        \EndIf
\EndWhile
\Return {\bfseries \itshape L$_{drift}$}
\end{algorithmic}
\end{algorithm}

\section{Experimental Analysis}
In this section, we empirically compare the proposed
method with other state-of-the-art (SOTA) methods for chunk data streams with various concept drifts. The effectiveness of the adaptive threshold setting method in
CADM and how it changes with cosine similarity are also illustrated.

\subsection{Experimental Settings}
Four simulation datasets with decision boundaries of different shapes are selected, which generally contain two variables .The extreme concept drift, i.e. labels of all samples are reversed, is simply introduced (see Fig. 3, 4).
Several advanced methods (DWM \cite{DWM}, ARF \cite{ARF}, HT \cite{HT}, NB \cite{NB}, CPSSDS \cite{CPSSDS}, OSELM \cite{OSELM}, BLS \cite{BLS}) are selected to be compared, which are implemented by {\itshape skmultiflow}\footnote[1]{https://scikit-multiflow.github.io/}. 

For the parameters of CADM, the size of the window is set at 10, the label ratio is set at 0.2, and the size of the chunk is 200. We set $k$ as 2 by analogy to the probability criterion of Gaussian distribution\footnote[2]{The code is available at https://github.com/songqiaohu/CADM-Confusion-Model}.
\begin{figure}[!ht]
\centering
\begin{minipage}[c]{0.2\linewidth}
    \centering
    \includegraphics[width=\textwidth]{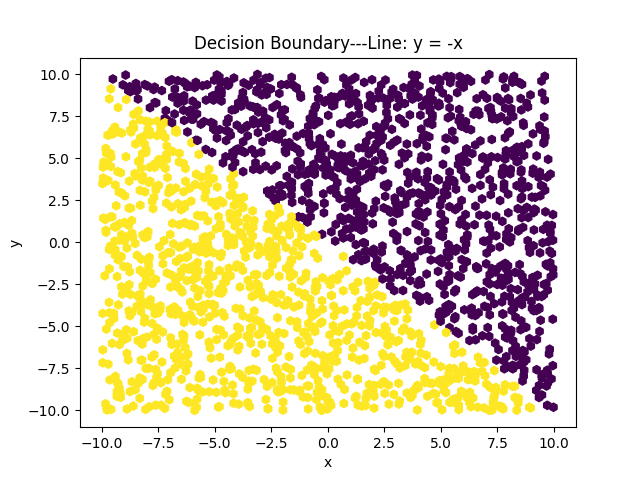}
    \centerline{(a) Line}
\end{minipage}%
\begin{minipage}[c]{0.2\linewidth}
    \centering
    \includegraphics[width=\textwidth]{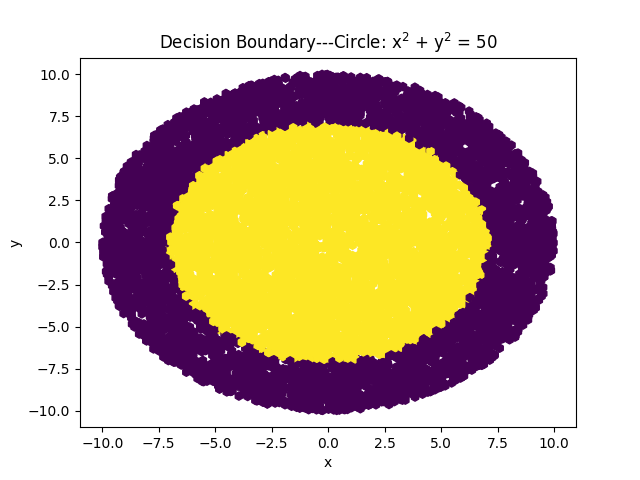}
    \centerline{(b) Circle}
\end{minipage}
\begin{minipage}[c]{0.2\linewidth}
    \centering
    \includegraphics[width=\textwidth]{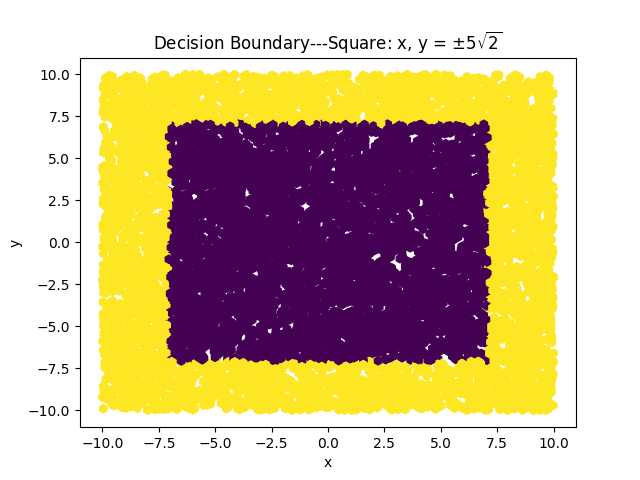}
    \centerline{(c) Square}
\end{minipage}
\begin{minipage}[c]{0.2\linewidth}
    \centering
    \includegraphics[width=\textwidth]{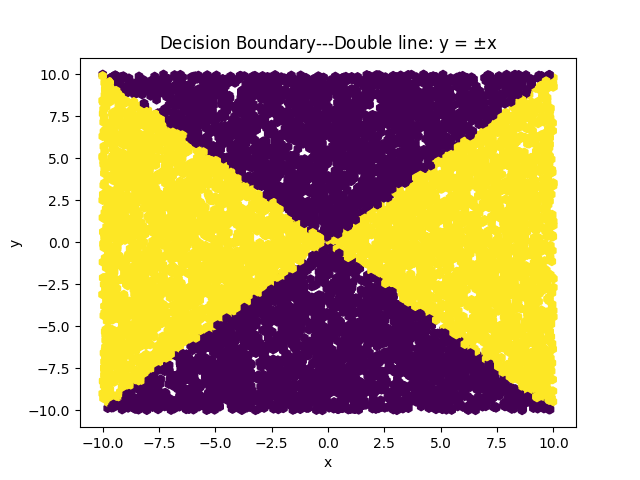}
    \centerline{(d) Double lines}
\end{minipage}
\caption{Simulation datasets with different shape-decision boundaries.}
\end{figure}

\begin{figure}[!ht]
\centering
\includegraphics[width=0.8\textwidth]{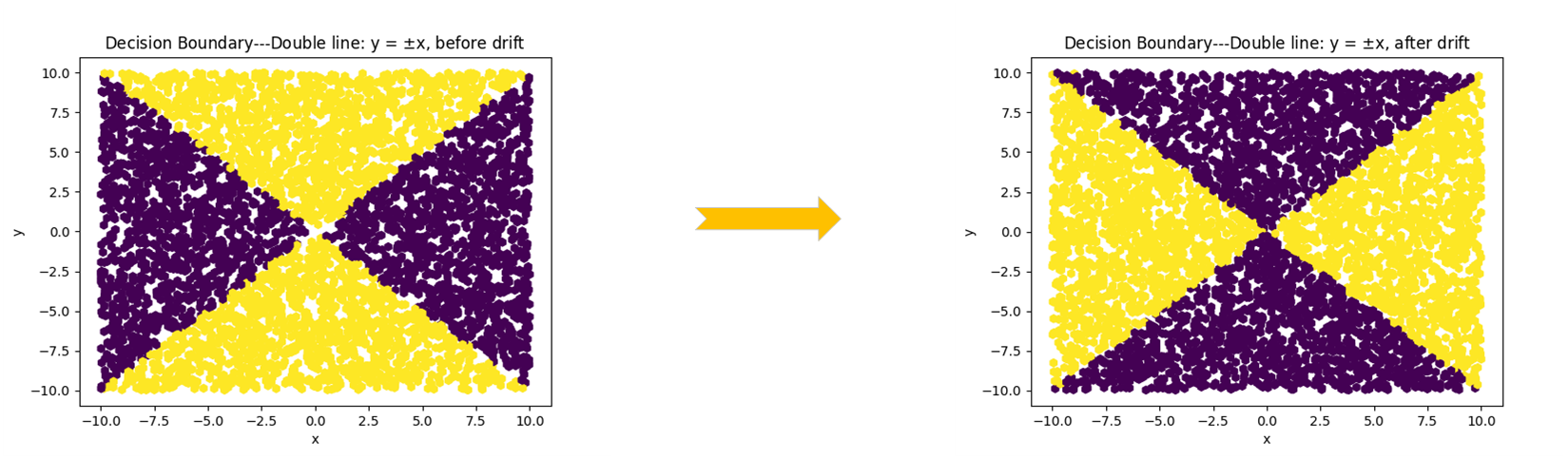}
\caption{Distribution of samples and labels before and after drifts.}
\end{figure}
        
\subsection{Experimental Results}
\subsubsection{Drift Detection and Threshold Change}
 In four simulation datasets, we set an extreme real drift at every 25 chunks, respectively. Naive Bayes [11] is used in CADM for this experiment, where the default parameters provided by
{\itshape scikit-multiflow} library in Python are considered. The change in cosine similarity and threshold are shown in Fig. 5. The effect of drift detection is reported in Table. 1. Clearly, {\itshape drift chunk} refers to the chunk where drift begins, {\itshape FA} refers to {\itshape False Alarm}, and the table content is the position where drift is detected.

 \begin{figure}[!ht]
 \centering
 \begin{minipage}[c]{0.4\linewidth}
     \centering
     \includegraphics[width=\textwidth]{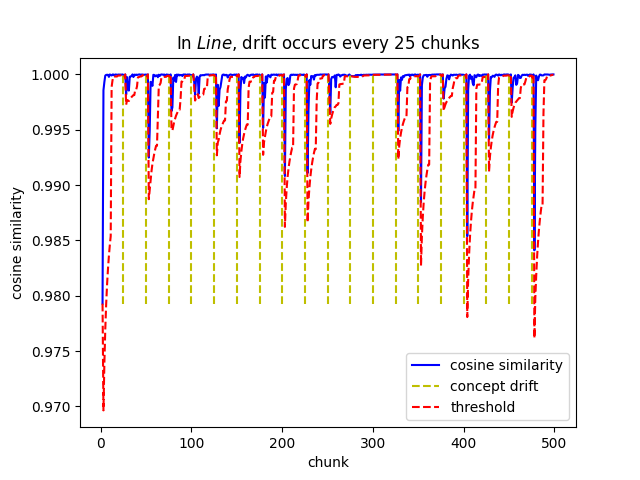}
     \centerline{(a) Line}
 \end{minipage}%
 \begin{minipage}[c]{0.4\linewidth}
     \centering
     \includegraphics[width=\textwidth]{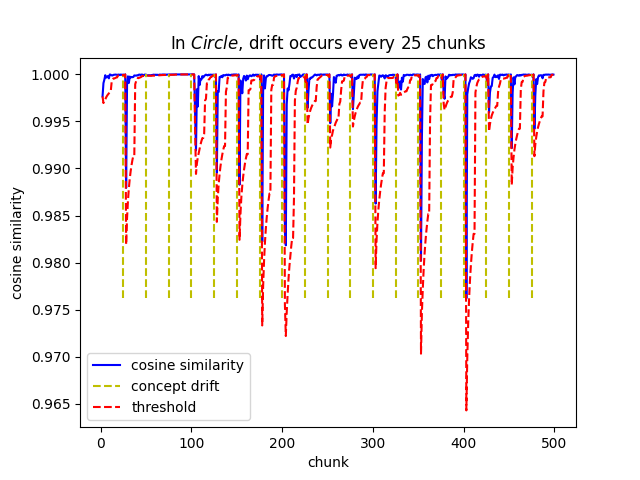}
     \centerline{(b) Circle}
 \end{minipage}
 \begin{minipage}[c]{0.4\linewidth}
     \centering
     \includegraphics[width=\textwidth]{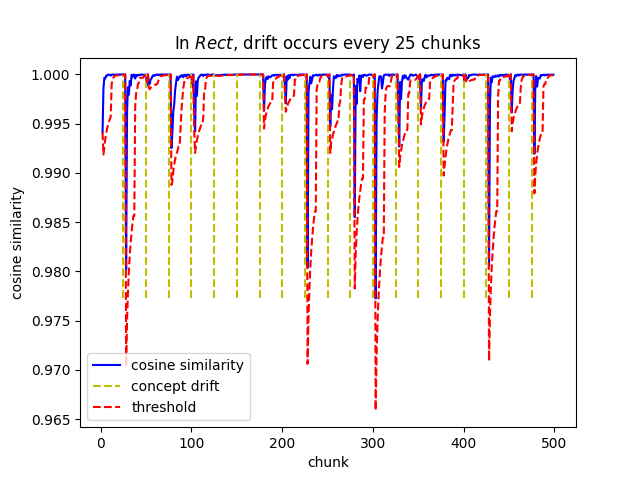}
     \centerline{(c) Square}
 \end{minipage}
 \begin{minipage}[c]{0.4\linewidth}
     \centering
     \includegraphics[width=\textwidth]{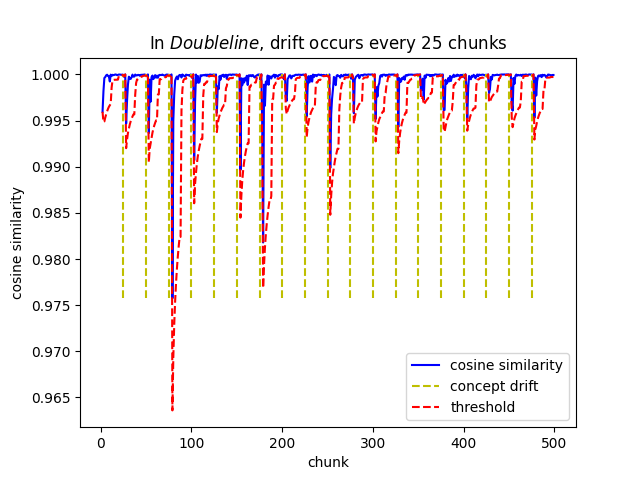}
     \centerline{(d) Double lines}
 \end{minipage}
 \caption{Cosine similarity and threshold varies with chunk.}
 \end{figure}

 \begin{table}
    \centering
    \fontsize{6}{12}\selectfont    
    \caption{Drift detection of CADM in different datasets.}
\begin{tabular}{c|cccccccccccccccccccc}
    \toprule
\diagbox [width=10em,trim=l] {Dataset}{Drift chunk} & 26 & 51 & 76  & 101 & 126 & 151 & 176 & 201 & 226 & 251 & 276 & 301 & 326 & 351 & 376 & 401 & 426 & 451 & 476 & FA\\
\hline
$Line$ & 27 & 52 & 77 & 102 & 127 &152 & 178 & 202 & 227 & 251 & $\times$ & $\times$ & 327 & 352 & 377 & 403 & 427 & 452 & 477 & -\\
$Circle$ & 27 & $\times$ & $\times$ & 103 & 127 &152 & 177 & 202 & 227 & 252 & 277 & 302 & 327 & 352 & 377 & 402 & 427 & 452 & 477 & -\\
$Square$ & 27 & 52 & 77 & 102 & $\times$ & $\times$ & 179 & 202 & 227 & 252 & 279 & 302 & 327 & 352 & 377 & 402 & 427 & 452 & 477 & -\\
$Doubleline$ & 27 & 52 & 78 & 102 & 127 & 153 & 177 & 203 & 226 & 252 & 277 & 302 & 327 & 352 & 377 & 402 & 427 & 452 & 477 & -\\
\bottomrule
\end{tabular}\vspace{0cm}
\end{table}
    
From this experiment, it can be verified that the proposed method has great advantages against false alarms. However, each drift detection has a certain delay, which is caused by two reasons: one is the principle of the method itself. The model should need to be confused before drift can be detected; the other is the randomness of samples. More samples are needed if a small number of samples cannot sufficiently confuse the model.
\subsubsection{Prediction Accuracy}
In this experiment, we compare the performance of the previously mentioned algorithms and their combination with CADM in prediction accuracy. The adjustable parameters of all methods are identical, including training data, label ratio, etc. The dataset used is {\itshape Doubleline}, where the decision boundary is nonlinear.

    \begin{figure}[!ht]
    \centering
    \begin{minipage}[c]{0.32\linewidth}
        \centering
        \includegraphics[width=\textwidth]{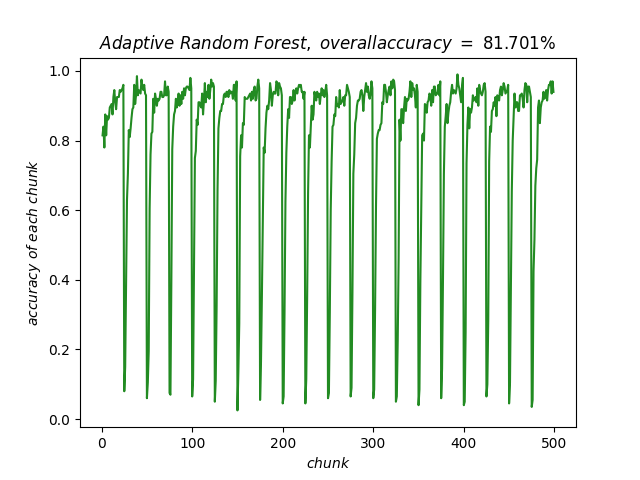}
        \centerline{(a) $ARF,81.701\%$}
    \end{minipage}%
    \begin{minipage}[c]{0.32\linewidth}
        \centering
        \includegraphics[width=\textwidth]{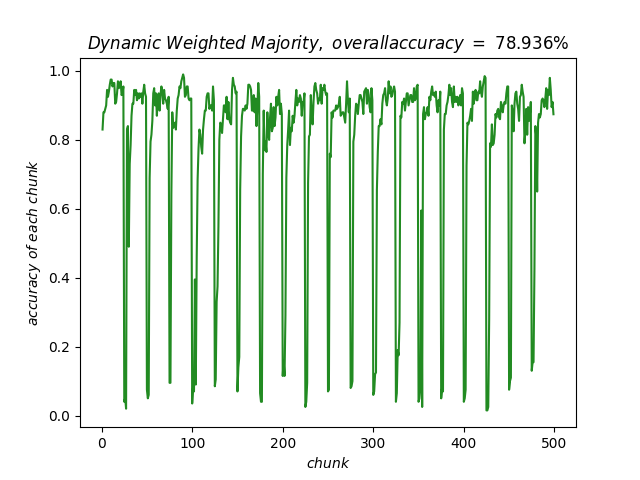}
        \centerline{(b) {\emph {DWM,}} $78.936\%$}
    \end{minipage}
    \begin{minipage}[c]{0.32\linewidth}
        \centering
        \includegraphics[width=\textwidth]{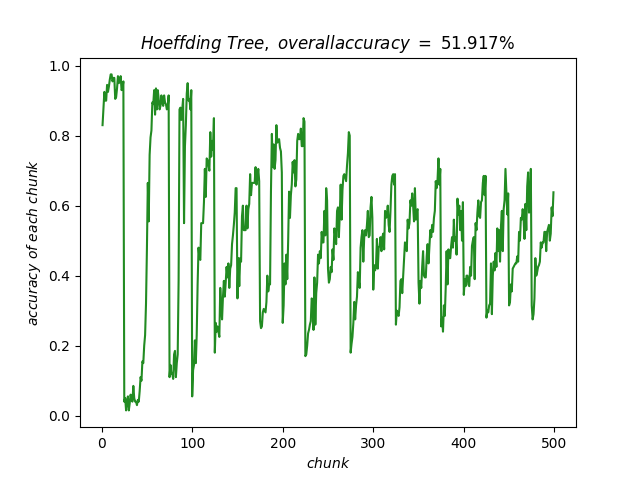}
        \centerline{(c) $CPSSDS,51.917\%$}
    \end{minipage}
    \begin{minipage}[c]{0.32\linewidth}
        \centering
        \includegraphics[width=\textwidth]{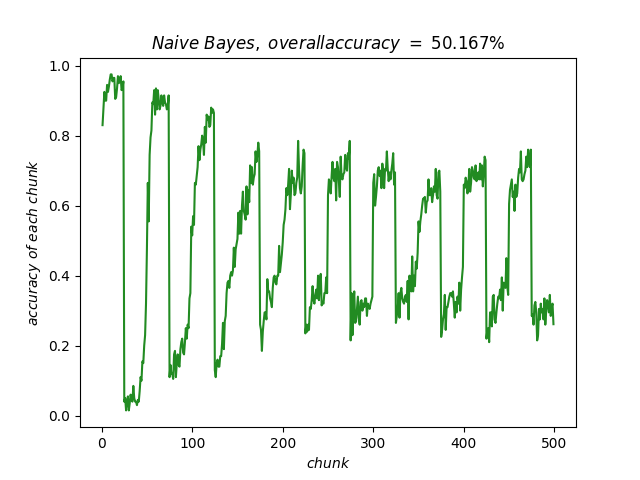}
        \centerline{(d) $NB,50.167\%$}
    \end{minipage}%
    \begin{minipage}[c]{0.32\linewidth}
        \centering
        \includegraphics[width=\textwidth]{HoeffdingTree.png}
        \centerline{(e) $HT,51.917\%$}
    \end{minipage}
    \begin{minipage}[c]{0.32\linewidth}
        \centering
        \includegraphics[width=\textwidth]{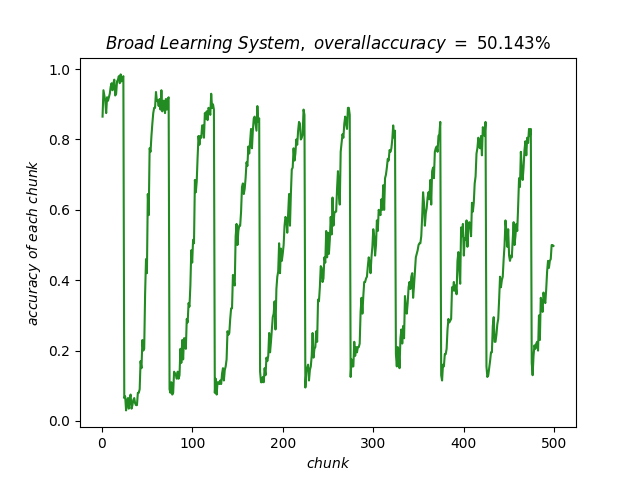}
        \centerline{(f) $BLS,50.143\%$}
    \end{minipage}
    \begin{minipage}[c]{0.32\linewidth}
        \centering
        \includegraphics[width=\textwidth]{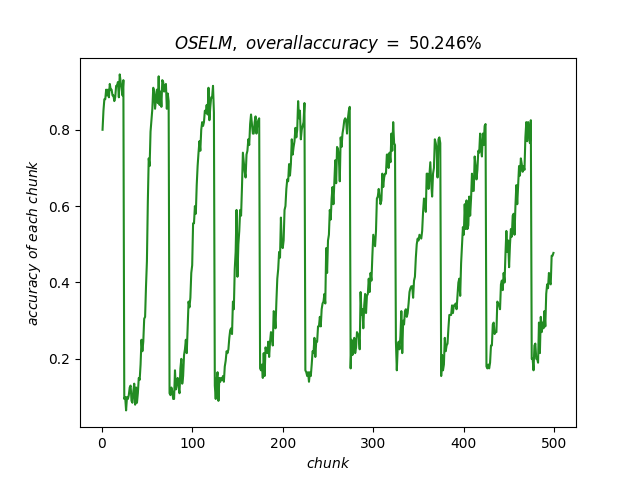}
        \centerline{(g) $OSELM,50.246\%$}
    \end{minipage}
    \begin{minipage}[c]{0.32\linewidth}
        \centering
        \includegraphics[width=\textwidth]{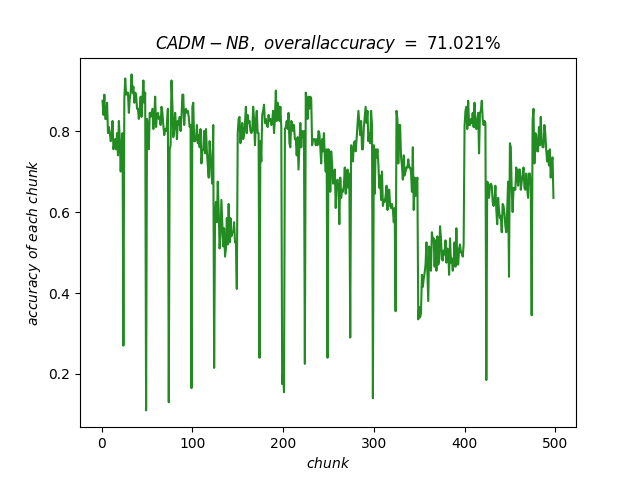}
        \centerline{(h) \emph{CADM-NB,} $71.021\%$}
    \end{minipage}
    \begin{minipage}[c]{0.32\linewidth}
        \centering
        \includegraphics[width=\textwidth]{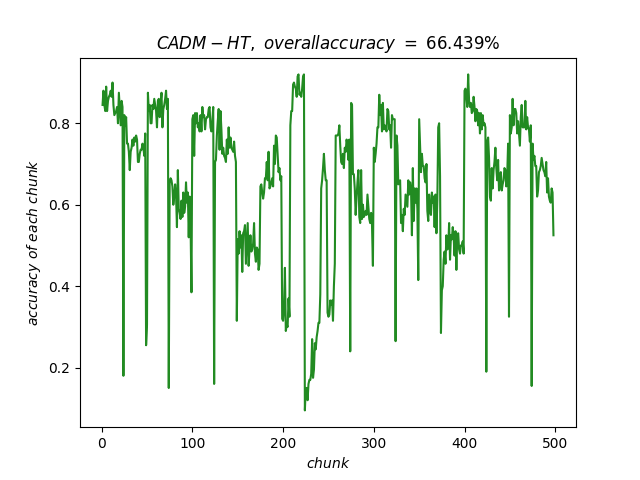}
        \centerline{(i) \emph{CADM-HT,} $66.439\%$}
    \end{minipage} 
    \begin{minipage}[c]{0.32\linewidth}
        \centering
        \includegraphics[width=\textwidth]{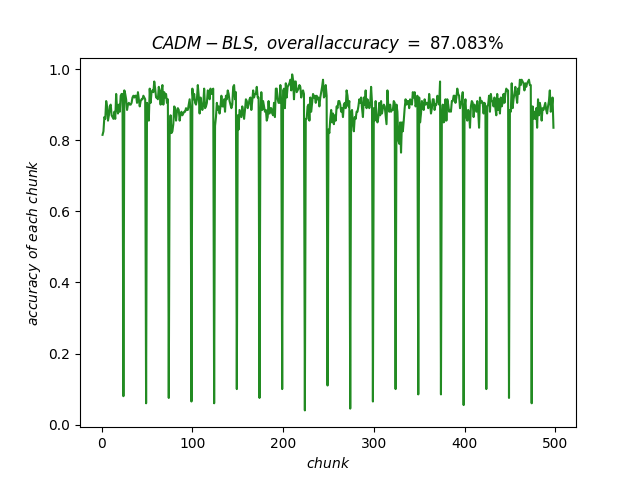}
        \centerline{(j) \emph{CADM-BLS,} $87.083\%$}
    \end{minipage}
    \begin{minipage}[c]{0.32\linewidth}
        \centering
        \includegraphics[width=\textwidth]{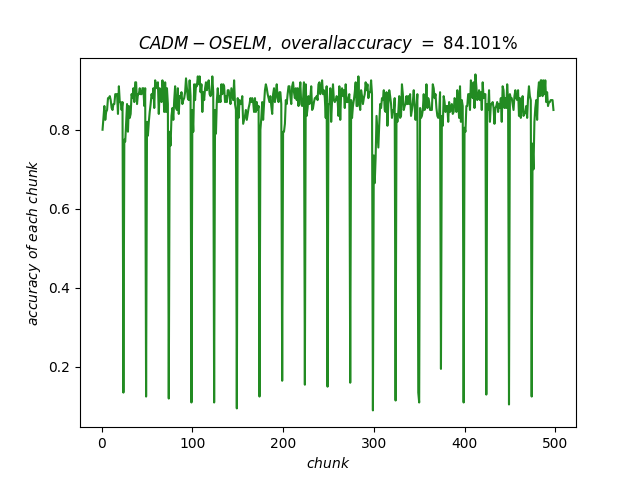}
        \centerline{(k) \emph{CADM-OSELM,} $84.101\%$}
    \end{minipage}
    \caption{Prediction accuracy of different methods.}
    \end{figure}

\begin{table}
    \centering
    \fontsize{6}{12}\selectfont    
    \caption{Overall accuracy for all methods in {\itshape Doubleline} (over ten runs).}
\begin{tabular}{c|c|c}
    \toprule
\multicolumn{2}{c|}{Methods} & Accuracy $\pm$ std(\%)\\
\hline
\multirow{3}*{Methods with drift detection} & $ARF$ & {\bfseries 81.164 $\pm$ 0.215 }\\
 & $DWM$ & {\bfseries 78.973 $\pm$ 0.872}\\
 & $CPSSDS$ & 51.917 $\pm$ 0.000\\
\hline
\multirow{4}*{Methods without drift detection} & $NB$ & 49.846 $\pm$ 0.297\\
 & $HT$ & 51.969 $\pm$ 0.590\\
 & $BLS$ & 49.780 $\pm$ 0.134\\
 & $OSELM$ & 49.801 $\pm$ 0.193\\
 \hline
 \multirow{4}*{Methods with $CADM$} & $CADM-NB$ & 65.773 $\pm$ 4.338\\
 & $CADM-HT$ & 64.893 $\pm$ 3.881\\
 & $CADM-BLS$ & {\bfseries 86.851 $\pm$ 1.156}\\
 & $CADM-OSELM$ & {\bfseries 79.293 $\pm$ 5.780}\\
\bottomrule
\end{tabular}\vspace{0cm}
\end{table}

From Fig. 6 and Table. 2, the following conclusions can be drawn: {\bfseries \romannumeral1)}Drifts in simulation datasets frequently occur. Without drift detection, the updating and adaptation speed of {\itshape NB, HT, BLS} and {\itshape OSELM} is slower than that of drift change, especially in the environment with few labels. Therefore, the accuracy curve almost changes following the drifts as shown in Fig. 6 (a)-(d); {\bfseries \romannumeral2)}After combining with {\itshape CADM}, all methods can quickly detect and adapt to the drifts. Hence, the accuracy curve can rapidly rise to a high position after a sharp fall. The overall accuracy improves significantly, as shown in Fig. 6 (e)-(h), which also shows the scalability and applicability of {\itshape CADM}; {\bfseries \romannumeral3)}Among all methods, {\itshape CADM-BLS} achieves the highest accuracy, which is higher than {\itshape ARF} and {\itshape DWM}. {\itshape CADM-OSELM} is also higher than {\itshape DWM}, which shows that the combinations of {\itshape CADM} and some base classifiers outperform the current mainstream algorithms with drift detection in {\itshape scikit-multiflow}. {\bfseries \romannumeral4)} For {\itshape CPSSDS} shown in Fig. 6 (k), its performance is similar to the methods without drift detection. As mentioned in the first section, it can only detect virtual drift rather than real drift.

\section{Conclusion}
In this paper, we have proposed the {\itshape CADM} to deal with real concept drift with limited annotations. {\itshape CADM} updates the model with manually annotated samples and pseudo-label-samples predicted by the model simultaneously, which can improve the performance of the model when there is no drift and cause confusion when there is drift. Then, cosine similarity and adaptive threshold have also been proposed to judge whether drift occurs. We have verified its effectiveness and superiority in extreme-drifting simulation datasets by combining CADM with different classifiers.\\
\\
\\
\\
\\
{\textbf{Acknowledgment.}} This work was supported by National Natural Science Foundation of China under Grant 61733009, National Key Research and Development Program of China under Grant 2017YFA0700300, and Huaneng Group science and technology research project.


%
%
%
%

\end{document}